\documentclass[10pt, a4paper]{article}

\usepackage{times}
\usepackage[utf8]{inputenc}
\usepackage[T1]{fontenc}
\usepackage{graphicx}
\usepackage{hyperref}
\usepackage{multirow}
\usepackage{multicol}
\usepackage{makecell}

\usepackage{lrec-coling2024} 

\name{Danrun Cao$^{1, 2}$, Nicolas Béchet$^{1}$, Pierre-François Marteau$^{1}$} 

\address{(1) Univ. Bretagne Sud, CNRS, IRISA \\
            Rue Yves Mainguy, 56000 Vannes, France \\
         \{danrun.cao, nicolas.bechet, pierre-francois.marteau\}@univ-ubs.fr\\
         (2) OctopusMind\\
         2 Pl. Saint-Pierre, 44000 Nantes, France \\
         d.cao@octopusmind.info\\}

\title{WikiNER-fr-gold: A Gold-Standard NER Corpus}

\abstract{
  We address in this article the the quality of the WikiNER corpus, a multilingual Named Entity Recognition corpus, and provide a consolidated version of it. 
  The annotation of WikiNER was produced in a semi-supervised manner i.e. no manual verification has been carried out \textit{a posteriori}. Such corpus is called silver-standard. 
  In this paper we propose WikiNER-fr-gold which is a revised version of the French proportion of WikiNER. 
  Our corpus consists of randomly sampled 20\% of the original French sub-corpus (26,818 sentences with 700k tokens). 
  We start by summarizing the entity types included in each category in order to define an annotation guideline, and then we proceed to revise the corpus. 
  Finally we present an analysis of errors and inconsistency observed in the WikiNER-fr corpus, and we discuss potential future work directions.
   \\ \newline 
}

\begin{document}
\maketitleabstract

\section{Introduction}
In Natural Language Processing (NLP), Named Entity Recognition (NER) is a task that focuses on identifying entities within unstructured text. 
The goal of an NER system is to locate nominal phrases referring to an entity and assign them a category from a predefined list. 
This phrase is referred to as the mention of an entity, and defined as a series of one or multiple consecutive tokens corresponding to one specific and unique entity. 
A token is defined as a continuous sequence of non-empty characters, representing the minimal unit during the automatic processing of textual data.
The NER task has a dual objective: determining the boundaries of a mention and categorizing the entity that is mentioned.

Training an NER system requires an annotated corpus. 
For French language there exists annotated corpora, but few are freely available. 
The French Treebank \cite{abeille_building_2003}, composed of articles from the newspaper \textit{Le Monde} (1990-1993), is a corpus for French syntactic analysis. 
It served as the basis for one of the first corpora dedicated to French NER, as presented in \cite{sagot_annotation_2012}. 
This corpus consists of 5,890 sentences with a total of 11,636 entities. 
Another usable corpus is Europeana Newspapers \cite{neudecker_open_2016}, which contains digitized newspaper articles processed with OCR tools. 
It is a multilingual corpus, and the French part contains 12,551 sentences. 
However, this corpus requires significant correction work before use, because many OCR-related errors remain disseminated in the corpus. 
The FENEC corpus \cite{millour_fenec_2022, millour_unveiling_2024} was created from six text genres (prose, poetry, journalistic text, encyclopedia, speech, and multi-sources). 
This corpus contains 11,149 tokens and 875 entities and was annotated following the Quaero schema \cite{rosset_entites_2011}.

The largest NER corpus we have identified is WikiNER \cite{nothman_learning_2013}, an encyclopedic corpus covering ten languages, including French. 
Several open-source NER tools have been trained on this corpus, such as \hyperlink{https://spacy.io/}{spaCy}, Flair \cite{akbik_flair_2019}, and \hyperlink{https://nlp.johnsnowlabs.com/}{Spark NLP}. 
This corpus consists of sentences extracted from Wikipedia articles, annotated with named entities. 
It covers four types of entities: person (PER), location (LOC), organization (ORG), and miscellaneous (MISC). 
All ten sub-corpora have the same size, comprising approximately 3.5 million tokens, making it a very substantial dataset. 
The annotations were produced in a semi-supervised manner, and there was no manual verification for the corpus. 
Therefore, it is considered a silver standard corpus.

In this article, we describe the manual correction process implemented to create a gold standard version of WikiNER. 
We will refer to this new corpus as WikiNER-fr-gold in the following discussion.
This work involved manual correction of 20\% of the French portion of WikiNER, which we will refer to as WikiNER-fr. 
WikiNER-fr-gold comprises 26,818 sentences and approximately 700,000 tokens.  
These data were randomly selected from the original corpus.
The revised corpus has been made available on \href{https://huggingface.co/datasets/danrun/WikiNER-fr-gold}{HuggingFace}.

The paper is organized as follows.
In \hyperref[sec:wikiner]{section 2}, we will provide a brief overview of the production of WikiNER annotations to highlight the origin of typical errors. 
\hyperref[sec:correction]{Section 3} will present the observed errors along with the correction choices we have made. 
Finally, in \hyperref[sec:conclu]{section 4}, we will present the future works.

\section{Production of the original annotations of WikiNER}
\label{sec:wikiner} 
The original annotations of WikiNER were produced using hyperlinks of Wikipedia articles. 
If there exists a Wikipedia page corresponding to an entity mentioned in a sentence, then the phrase describing that object would be linked to its Wikipedia page via a hyperlink. 
This linkage can be exploited in a reverse way: the text of a hyperlink helps identify an object, which matches the definition of a named entity. 
The boundaries of the hyperlink naturally serve as those of the mention. 
It remains simply to project the category of the object onto the mention. 
Annotation of the original corpus was thus carried out in two steps: the classification of Wikipedia pages, and the annotation of mentions within Wikipedia articles.

Firstly, for each of the 10 languages, the authors created a training corpus to train a classification model. 
The French corpus consists of approximately 2,500 articles. 
The annotations follow an extended version of the annotation schema of the BBN corpus \cite{brunstein_annotation_2002}. 
Next, the authors compared three classification strategies to find the best solution. 
A classifier was trained using each of these strategies, and the authors reported precision, recall, and F1 score using 10-fold cross-validation. 
The best method was Logistic Regression with an average F1 score of 94\%. 
It was then used for the classification of the remaining Wikipedia pages.

Then, the categories of Wikipedia pages were projected onto their hyperlinks occurring in other Wikipedia pages in order to make the initial annotations.
In Wikipedia, only the first occurrence of an entity receives a hyperlink.
The authors proposed several inference strategies in order to retrieve the other mentions in the remaining text.
First a list of potential mentions was created for each entity.
This list is generated from hyperlinks and redirections to corresponding Wikipedia pages.
Not every element of this list is eligible as a mentioned candidate, obviously.
The authors then proposed several criteria to filter non-conforming elements.
Four rigor levels are defined by varying criteria combinations.
A higher level of rigor corresponds to a more strict filtration.
Annotation quality may be higher but at the cost of a reduced variety level of mentions.
In total, five variants of the corpus are proposed, each with 3.5 million tokens.

In our study, we chose WIKI-2, the version produced with level-2 filtration. 
It represents a good compromise between annotation quality and entity coverage. 
Table \ref{tab:dist_tag} displays the token count for each entity type in the corpus. 
Each token can only receive a single label as its entity type and can belong to only one entity.

\begin{table}
    \centering
    \begin{tabular}{c|cccc}
         Entity & PER & LOC & ORG & MISC \\
         \hline
         \makecell{Token\\count} & 129,978 & 155,565 & 45,443 & 81,594
    \end{tabular}
    \caption{Token count of each entity type in WIKI-2}
    \label{tab:dist_tag}
\end{table}

\section{Corpus review}
\label{sec:correction}
\subsection{Entity category definition}

The annotation scheme serves to clarify how the categories were defined. 
We propose summarizing this annotation schema by presenting the types of entities included in each category. 
Table \ref{tab:entity_type} provides a comprehensive list of entity types by category, with reference to some examples.

\begin{table*}[t]
\centering
  \label{tab:entity_type}
  \begin{tabular}{|c|c|c|}
  \hline
  Category & Entity type & Example \\
    \hline
    \multirow{5}{*}{\textbf{LOC}} & Country and region & \textit{France}, \textit{Loire Atlantique}\\
    \cline{2-3} & Iconic building & \textit{Gare Montparnasse}, \textit{Tour Eiffel} \\
    \cline{2-3} & Natural landscape & \textit{Seine}, \textit{Alpes} \\
    \cline{2-3} & Transport lines and networks & \textit{TGV Est}, \textit{RER A} \\
    \cline{2-3} & Celestial bodies & \textit{Soleil (Sun)}, \textit{Alpha Centauri} \\
    \hline
    \multirow{3}{*}{\textbf{PER}} & Name and family & \textit{Staline}, \textit{Maison d'Orange}\\
    \cline{2-3} & Fictional characters & \textit{Zeus}, \textit{Indiana Jones} \\
    \cline{2-3} & Nationality and ethnicity & \makecell{\textit{(les) Français (the French)},\\ \textit{(les) Aztèques (the Aztecs)}} \\
    \hline
    \multirow{11}{*}{\textbf{ORG}} & Organization, institution & \makecell{\textit{ONU (United Nations)}, \\ \textit{Fonds monétaire international} \\ \textit{(International Monetary Fund)}}\\
    \cline{2-3} & Government bodies & \makecell{\textit{Assemblée Générale (General Assembly)},\\\textit{Parlement Irlandais (Irish Parliament)}} \\
    \cline{2-3} & Political parties & \makecell{\textit{UMP}, \textit{Parti communiste chinois} \\ \textit{(Chinese Communist party)}} \\
    \cline{2-3} & Companies & \textit{Microsoft}, \textit{EDF (Electricity of France)} \\
    \cline{2-3} & Sports teams & \makecell{\textit{Bulls de Chicago (Chicago Bulls)}, \\ \textit{(équipe) France (French national team)}} \\
    \cline{2-3} & Musical bands & \textit{les Beatles}, \textit{AC/DC} \\
    \cline{2-3} & \makecell{Higher education institutions} & \makecell{\textit{Université de Paris}, \\ \textit{Université de Californie à Berkeley} \\ \textit{(UC Berkeley)}} \\
    \cline{2-3} & Military organizations & \textit{Armée Rouge (Red Army)}, \textit{US Marine Corps} \\
    \hline
    \multirow{8}{*}{\textbf{MISC}} & Titles of works & \textit{La Joconde (Mona Lisa)}, \textit{Bible}\\
    \cline{2-3} & Events & \makecell{\textit{Seconde Guerre Mondiale} \\ \textit{(World War II)},\\\textit{Jeux Olympiques (Olympic Games)}} \\
    \cline{2-3} & Historical periods and regimes & \makecell{\textit{Dynastie Qing (Qing Dynasty)},\\ \textit{Grèce antique (Ancient Greece)}} \\
    \cline{2-3} & \makecell{Software and hardware} & \textit{(langage) Python}, \textit{PS5} \\
    \cline{2-3} & Conventions and documents & \makecell{\textit{Édit de Nantes (Edict of Nantes)},\\ \textit{(la) Constitution}} \\
    \cline{2-3} & Ships and rockets & \textit{HMS Triumph}, \textit{Ariane 2} \\
    \cline{2-3} & Brands & \textit{Land Rover}, \textit{TGV} \\
    \hline
\end{tabular}
\caption{Entity types by category with examples}
\end{table*}

\subsection{Annotation format and tool}
The annotations are formatted in BIOES format. 
Within each entity, we distinguish the beginning (B), inside (I), and end (E) of the entity. 
This format helps highlight the boundaries of entities.
For example, in "\textit{général de Gaulle (General de Gaulle)}", the three tokens receive the labels B-PER, I-PER, and E-PER, respectively. 
For entities consisting of a single word, we use the label S (for single). 
So, the entity "France" is annotated as S-LOC. 
Tokens outside entities are labeled as O, indicating they are not part of any entity. 
There are a total of 17 formatted labels.

\begin{figure*}[ht] 
\begin{center} 
\includegraphics[width=\textwidth]{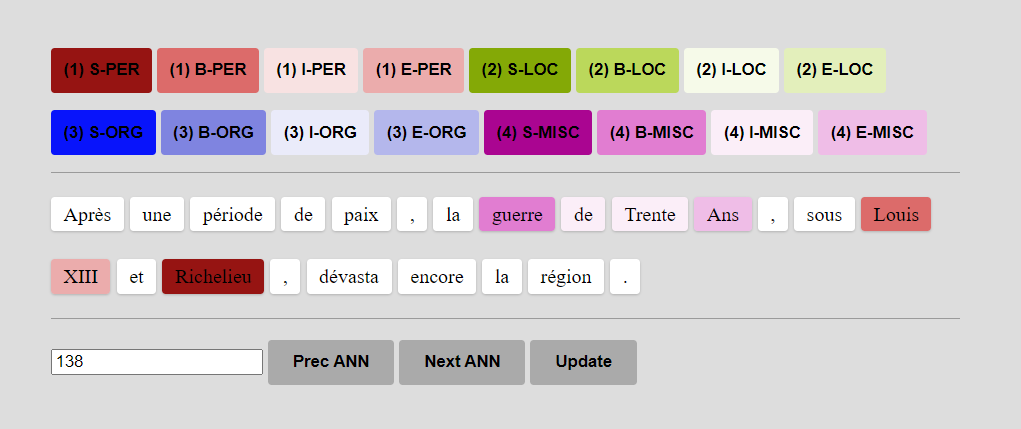}
\end{center} 
\caption{Overview of the labeling tool} \label{apercu} \
\end{figure*}

We use the labeling tool provided by (anonymous reference).
The advantage of this tool lies in its ability to customize candidate labels and their visual representation. 
Thanks to this, a color scheme could be defined that facilitates the understanding of the category and boundaries of the entity. 
Figure \ref{apercu} provides an overview of the tool's interface.

\subsection{Error analysis and correction}
During the corpus review, we observed very few clear-cut errors, meaning mentions that do not correspond to either an entity or a Wikipedia page. 
Most errors are recurring, and we can easily trace their origins in the annotation generation process. 
In the following paragraphs, we present these errors grouped by their nature. 
We then explain the corrections made accompanied by examples.

We would like to insist on the fact that the objective of our work is to solely standardize annotations and correct errors. 
We do not question the logic of the original annotation choices. 
Thus, as a principle, we do not change the category assigned to an entity unless it is an indisputable error (for example, annotating "\textit{France}" as MISC). 
In cases of incoherence, when an entity receives multiple categories, we refer to the annotation of other entities of the same type and to their corresponding Wikipedia articles to decide whether or not a modification should be made. 
Also, it is important to note that this review is only applied to entities that have already been identified. 
We do not add entities unless there is an obvious omission, such as a country name that wasn't annotated.

\subsubsection{Inconsistent definition of hyperlinks}
\label{sec:type1}
The hyperlinks in Wikipedia are manually created by many contributors. 
There may be a lack of agreement on hyperlink standards, which can result in the generation of inconsistent annotations. 
For example, in the phrase "\textit{la France (France)}", some link the word "\textit{France}" to the corresponding page, while others also include the article "\textit{la}" in the mention. 
As a result, in the corpus, both the mentions "\textit{France}" and "\textit{la France}" exist for the same entity "\textit{France}". 
Similarly, appositions can lead to inconsistencies. 
The mention "\textit{ville de Lyon (city of Lyon)}" was seen associated with the entity "\textit{Lyon}" instead of the token "\textit{Lyon}" by itself.

Aside from redundant mentions, there exist also incomplete mentions. 
For example, in the mention "\textit{Coupe du monde (World Cup)}", sometimes only the word "\textit{Coupe (Cup)}" is annotated. 
This phenomenon is especially common with nested entities (entities that contain other entities). 
Take the example of "\textit{comté de Mortain (County of Mortain)}", a medieval county centered around the town of Mortain. 
The entire mention should receive the LOC label, but instead only the town "\textit{Mortain}" has been annotated.

For this type of error, it will suffice to simply remove redundant parts and add missing ones. 
As a general rule, articles, appositions, and descriptions are removed from the mention, except in two cases.
The first case is when they are part of the entity's name or its conventional appellation. 
For example, "\textit{Le Havre}" remains "\textit{Le Havre}". 
"\textit{Général de Gaulle (General de Gaulle)}" also remains a complete mention, although "\textit{général (general)}" is not part of the name. 
On the contrary, "\textit{le roi Louis XIV (King Louis XIV)}" becomes "\textit{Louis XIV}" because it is understood that he is a king without specifying it.
The second case is when their presence is essential to avoid ambiguity. 
Consider the pair "\textit{ville de Bruxelles (city of Brussels)}" and "\textit{Région de Bruxelles (Region of Brussels)}". 
The first refers to the city of Brussels, while the latter refers to a region in Belgium of which Brussels is the capital. 
This also applies in the PER (Person) category, such as "\textit{de Saint-André}," which refers to the journalist Alix de Saint-André, and "\textit{maréchal de Saint-André (Marshal de Saint-André)}", corresponding to the marshal Jacques d'Albon de Saint-André.

It could be difficult to differentiate a nested entity from a false annotation comprising the entity's description. 
In cases of doubt, we refer to Wikipedia for clarification. 
We first check the entity with the widest scope to see if there is a Wikipedia page with a title corresponding to the complete mention. 
If such a page exists, we retain that entity and annotate it. 
If not, we reduce the mention to the smaller entity and repeat the verification process. 
There are instances where nested entities have only certain components annotated. 
In these cases, we try to complete them as much as possible.

\subsubsection{Hyperlinks non conforming to the definition of a named entity}
Two criteria must be fulfilled for a phrase to be recognized as a mention of a named entity: (1) it must be of \textbf{nominal} nature and (2) it must refer to a \textbf{specific} and \textbf{unique} real-world object. 
However, Wikipedia pages and hyperlinks do not have such rules. 
For example, there exists a page "\textit{Relations entre la Chine et le Tibet durant la dynastie Ming (Relations between China and Tibet during the Ming Dynasty)}".
But it is not considered an entity since the relationship between two regions is not a clear and precise concept. 
A hyperlink leading to this page was placed on the phrase "\textit{La Dynastie Ming patronnait l’activité religieuse du Tibet (The Ming Dynasty sponsored religious activity in Tibet)}". 
The page was annotated as MISC, hence the phrase inherited the same annotation. 
Aside from the false annotation of the Wikipedia page, a complete phrase cannot be considered a named entity. 
So we remove the annotation on the phrase, and annotate only the entities "\textit{Dynastie Ming (Ming Dynasty)}" and "\textit{Tibet}". 
Similarly, "\textit{histoire de la Chine (history of China)}" and "\textit{liste de communes de France (list of municipalities in France)}" are not considered named entities.

Another peculiarity of named entities is that their interpretation depends on context. 
For example, in a general context, "\textit{Cité Interdite (Forbidden City)}" refers to the ancient royal palace in China.
Thus it is annotated as LOC. 
However, there is also a page about a film bearing the same name. 
Mentions related to this sense should be annotated as MISC. 
Furthermore, some entities cannot be interpreted without context. 
In the sentence "\textit{Sa mère meurt d’un cancer de l’estomac le 15 septembre 1821 (Her mother died of stomach cancer on September 15, 1821)}" from the page "\textit{Charlotte Brontë}", "\textit{Sa mère (Her mother)}" receives a hyperlink to the page "\textit{Maria Brontë}".
"\textit{Sa mère}" is therefore annotated as PER. 
However, this inference is valid only within the original context, i.e., in the article presenting Charlotte Brontë. 
Without this context, "\textit{sa mère}" cannot be associated with a specific person.
Therefore, this phrase is not considered a named entity, and its annotation is removed.

\subsubsection{Entities of complex nature}
\label{sec:type3}
Certain entities can be challenging to categorize due to their complex nature, especially geopolitical entities. 
For example, in Wikipedia, the "\textit{Empire Britannique (British Empire)}" is defined as "\textit{l’ensemble des territoires qui, sous des statuts divers [...] ont été gouvernés ou administrés du XVI au XX siècle par l’Angleterre, puis le Royaume-Uni}" (the set of territories that, under various statuses [...] were governed or administered from the 16th to the 20th century by England, then the United Kingdom). 
If we consider it as a group of colonies, then the entity can be seen as a geographical concept and annotated as LOC. 
However, there is also an organizational and hierarchical structure between the United Kingdom and its colonies. 
In this sense, it is also appropriate to annotate it as ORG. 
This discussion can apply to other entities of the same kind, such as "\textit{Empire romain (Roman Empire)}", "\textit{Grèce antique (Ancient Greece)}", and "\textit{Allemagne nazie (Nazi Germany)}".

Now consider "\textit{Carthaginois (Carthaginians)}" in the sentence "\textit{Les Carthaginois prennent d’abord la ville de Messine}" (The Carthaginians first take the city of Messina). 
Annotating it as PER seems right since it refers to the people of the Carthaginian civilization that occupied Messina. 
However, the entire population did not participate in the war, but rather the Carthaginian army. 
Following this logic, "\textit{Carthaginois}" should be annotated as ORG. 
But once again, army or people, war is an act that involves two nations. 
Therefore, it would also be possible to annotate this entity as LOC.

For such entities, it is difficult to assign a single label, and the annotation choices of the authors can be easily contested. 
We try to follow the original annotation schema when dealing with them. 
If the entity appears elsewhere in the corpus, we adopt the same label. 
If not, we refer to other entities of the same type and assign a label that we find most appropriate. 
One special case regards nationalities or ethnicities, such as "\textit{Carthaginois}". 
The annotation of these entities is highly diverse, all four labels can be found. 
We have made the decision to annotate them all as PER, but the debate remains open.

\section{Conclusion}
\label{sec:conclu}
We have presented WikiNER-fr-gold, a gold-standard NER corpus in French. 
The corpus consists of 20\% of the WikiNER-fr corpus, randomly sampled, which is then subjected to manual revision. 
Our goal was to standardize and homogenize the annotations while following the original annotation schema as much as possible. 

One limitation in this work is the lack of comparison with other annotation schemes.
For example, titles such as "\textit{Duc de Bretagne (Duke of Brittany)}" are considered an entity only when it refers to one specific person deductible from the sentential context.
This choice was made in coherence with the definition of a named entity.
However in the Quaero corpus, they are annotated PER, since a title is associated with a person, even when we do not know precisely which one.
It would have been interesting to compare the handling of such cases in other corpora, and if possible, to hear their authors' explanation on annotation choices.

In future works, we will perform a more comprehensive assessment of WikiNER's annotations regarding other NER corpora, with the goal of a revision of entity categorization. 
This could be the occasion, for example, to revisit the annotation of geopolitical entities. 
Ideally, this corrective process would be applied to the entire corpus. 
Some of the corrections can be automated, especially for certain recurring errors. 
Redundant articles, for instance, can be easily identified using rules and lexicons. 
We can also solicit the \hyperlink{https://www.mediawiki.org/wiki/API:Main_page}{Wikipedia API} to facilitate the detection of embedded entities. 
Furthermore, it would be interesting to implement an active learning system during the correction. 
We can train an assistant model that takes into account previously encountered errors and then identifies potential erroneous mentions. 
The new annotation guidelines will be distributed with the corpus to keep the task of expanding WikiNER-fr-gold open and active.
Finally, we will extend the revision work to the entire WikiNER-fr, and eventually to other languages.

\bibliographystyle{lrec-coling2024-natbib}
\bibliography{augmentation}

\end{document}